\documentclass[10pt,twocolumn,letterpaper]{article}

\usepackage{iccv}
\usepackage{times}
\usepackage{epsfig}
\usepackage{graphicx}
\usepackage{amsmath}
\usepackage{multirow}
\usepackage{amssymb}
\usepackage{booktabs}
\usepackage{pifont}
\usepackage{soul}
\usepackage{color, colortbl}
\usepackage[dvipsnames]{xcolor}

\usepackage[pagebackref=true,breaklinks=true,colorlinks,bookmarks=false]{hyperref}

\iccvfinalcopy % *** Uncomment this line for the final submission

\def\iccvPaperID{5} % *** Enter the ICCV Paper ID here

\ificcvfinal\fi

\graphicspath{{figures/}{figures/nnr/}}
\DeclareGraphicsExtensions{.pdf,.png}

\newcommand{\fig}[1]{Figure \ref{#1}}
\newcommand{\eq}[1]{eqn. (\ref{#1})}

\newcommand{\sect}[1]{Section \ref{#1}}
\newcommand{\tab}[1]{Table \ref{#1}}

\newcommand{\myparagraph}[1]{\vspace{2pt}\noindent{\bf #1}}
\newcommand{\cmark}{\Huge\textcolor{green}{\ding{51}}}%
\newcommand{\xmark}{\Huge\textcolor{red}{\ding{55}}}%
\newcommand{\nCr}[2]{\,^{#1}C_{#2}} % nCr

\definecolor{babyblueeyes}{rgb}{0.63, 0.79, 0.95}

\begin{document}

%%%%%%%%% TITLE
\title{Concurrent Discrimination and Alignment for Self-Supervised Feature Learning}

\author{Anjan Dutta$^1$, Massimiliano Mancini$^2$, Zeynep Akata$^2$\\
$\,^1$University of Exeter, $\,^2$University of T\"{u}bingen}

% \author{Anjan Dutta\\
% University of Exeter\\
% Institution1 address\\
% {\tt\small a.dutta@exeter.ac.uk}
% For a paper whose authors are all at the same institution,
% omit the following lines up until the closing ``}''.
% Additional authors and addresses can be added with ``\and'',
% just like the second author.
% To save space, use either the email address or home page, not both
% \and
% Massimiliano Mancini\\
% University of T\"{u}bingen\\
% First line of institution2 address\\
% {\tt\small m.mancini@uni-tuebingen.de}
% \and
% Zeynep Akata\\
% University of T\"{u}bingen\\
% First line of institution2 address\\
% {\tt\small zeynep.akata@uni-tuebingen.de}
% }

\maketitle

% Abstract
\begin{abstract}
Existing self-supervised learning methods learn representation by means of pretext tasks which are either (1) discriminating that explicitly specify which features should be separated or (2) aligning that precisely indicate which features should be closed together, but ignore the fact how to jointly and principally define which features to be repelled and which ones to be attracted. In this work, we combine the positive aspects of the discriminating and aligning methods, and design a hybrid method that addresses the above issue. Our method explicitly specifies the repulsion and attraction mechanism respectively by discriminative predictive task and concurrently maximizing mutual information between paired views sharing redundant information. We qualitatively and quantitatively show that our proposed model learns better features that are more effective for the diverse downstream tasks ranging from classification to semantic segmentation. Our experiments on nine established benchmarks show that the proposed model consistently outperforms the existing state-of-the-art results of self-supervised and transfer learning protocol. Code can be found at \url{https://github.com/AnjanDutta/codial}.
\end{abstract}

% Introduction
\section{Introduction}
\label{sec:intro}
Supervised deep learning methods achieve a high performance only when trained on a large amount of labeled data gathered through expensive and error-prone annotation procedure. Therefore, depending on the training procedure and underlying dataset, supervised learning might yield features that mainly focus on the local statistics, and hence might end up learning spurious correlations. Since it is well known that global statistics possess better generalization capability \cite{Jenni2020GlobStat}, supervised learning often suffers from a poor generalization ability.
\begin{figure}[!ht]
\centering
\includegraphics[width=\columnwidth]{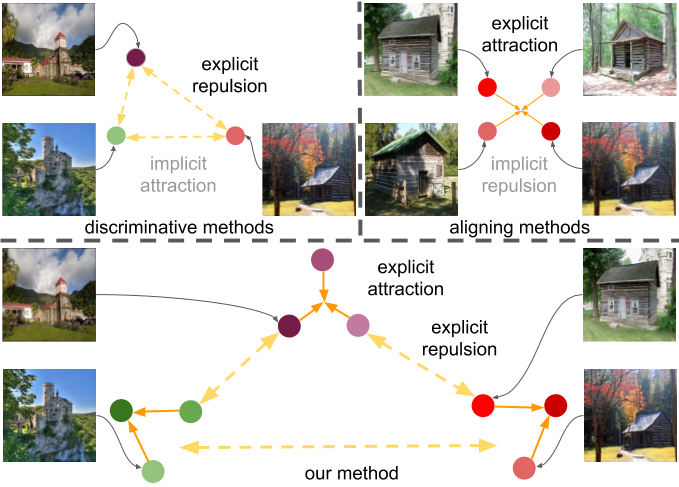}
\caption{Self-supervised learning methods are (1) discriminative, \ie explicitly specify which features should be repelled, and (2) aligning, \ie indicate which features should be attracted. Our method unifies the advantages of both the categories and explicitly formulates the features to be attracted and repelled simultaneously.}
\vspace{-15pt}
\label{fig:teaser_ssl}
\end{figure}
To alleviate these limitations \cite{Jing2020Survey}, and to learn representations from large-scale unlabeled datasets, self-supervised learning (SSL) \cite{Doersch2015Context,Doersch2017MultiTask,Oord2018CPC} has been proposed. One way to do that is to learn some discriminative tasks, such as the recognition of rotation angle \cite{Gidaris2018RotNet} or a local transformation \cite{Jenni2018SSLArtifacts,Jenni2020GlobStat} that needs understanding global statistics. Another strategy is to perform diverse image transformations for a single image, and then aligning them with some criterion \cite{Henaff2019CPC,Misra2020PIRL,Chen2020SimCLR,Grill2020BYOL,Fedirici2020MIB,Tian2020CMC}. In other words, the first group of methods specifies which features should be repelled from each other but does not describe explicitly which features should be attracted to each other (see Figure \ref{fig:teaser_ssl}, upper left). On the contrary, the second group specifies which features should be attracted, but does not set out which features should be repelled (see Figure \ref{fig:teaser_ssl}, upper right), which creates ambiguities in feature learning. Although \cite{Feng2019RotDecouple} proposes to recognize angle of image rotations and simultaneously minimizing the Euclidean distance between representations of the transformed images and their mean, this solution could yield trivial solution by simply learning the same zero vector as the image feature \cite{Ji2019IIC}.

%TODO: also motivate with shortcut solutions
To address this fundamental shortcoming of SSL methods, we propose CODIAL (COncurrent DIscrimintion and ALignment), a novel method that combines the benefit of discriminative and aligning methods. The pre-training task of our method is to discriminate the image transformations and concurrently align global image statistics. For doing that, we define primary and auxiliary image transformations which are orthogonal to each other and respectively use them to discriminate and align global image statistics. Motivated by \cite{Jenni2020GlobStat}, we choose the primary transformations in such a way that the local image statistics remain largely unchanged, such as recognition of image rotation angles, warping transformations. The auxiliary transformations are the standard ones for image augmentations, such as random crop, horizontal flip, color jittering, blurring etc \cite{Chen2020SimCLR,Grill2020BYOL}. The repulsion is achieved via a discriminative classifier and the feature alignment is done by engaging a Jensen-Shannon mutual information estimator \cite{Hjelm2019InfoMax,Fedirici2020MIB}.

In this work, we make the following contributions: (1) We introduce a novel multi-task method for self-supervised feature learning which combines the advantages of discriminative and alignment based works; (2) Our method successfully avoids degenerating and shortcut solutions by its design which we further enforce by introducing information bottleneck regularizer term to our learning objective; (3) We perform extensive experiments on nine benchmark datasets and achieve state-of-the-art performance both in self-supervised and transfer learning experiments.

% Related Work
\section{Related Work}
\label{sec:related}
In this section, we review the relevant literature of discriminative and alignment-based semi-supervised learning as well as hybrid methods.

\myparagraph{Discriminative Methods:} Majority of discriminative methods rely on using auxiliary handcrafted prediction tasks to learn their representation. Early works within this group try to determine spatial configuration of patches \cite{Doersch2015Context,Noroozi2016Jigsaw,Noroozi2018SSLKT,Mundhenk2018SSLContext}, which needs distinguishing local patches. Since local features are not sufficient to learn spatial configuration, these pretext tasks necessarily learn global statistics up to the size of local patch or tile. Balancing the local and global features has also been seen in obtaining and matching the local and global statistics \cite{Noroozi2018Count}. The pretext task of predicting rotation \cite{Gidaris2018RotNet} distinguishes the features of the images rotated by $0^\circ$, $90^\circ$, $180^\circ$, $270^\circ$, which essentially learns a representation suitable to recognize the angle. The pretext task involved in \cite{Jenni2018SSLArtifacts} uses adversarial training to distinguish real images from images with synthetic artifacts for learning visual representation. Few other works use pseudo labels obtained from an intermediate unsupervised clustering step for the networks \cite{Caron2018DeepClustering,Gansbeke2020SCAN}. Recently, Jenni \etal \cite{Jenni2020GlobStat} extend the work of Gidaris \etal \cite{Gidaris2018RotNet} and propose to discriminate local transformation together with some previously proposed predictive tasks. In this work, we use the rotation and warping as the primary transformations for using in the predictive task. Different to the prior works, we randomly crop the original image maintaining the specification mentioned in \sect{sec:expt} before generating each transformation, \ie we apply the rotation and warping transformations on the cropped version of the original image, which has been proven to be effective for our model.

%TODO: A brief overview on MI methods
\myparagraph{Alignment Methods:} Early works within this group apply some image transformations and align the transformed and real image for learning the underlying model. Inpainting \cite{Pathak2016ContextEncoders} and colorization \cite{Zhang2016Color,Zhang2017SplitBrain} are among the first few approaches that belong to this category. In case of inpainting, the pretext task involves removing a set of pixels as transformation and then reconstructing the missing part. The pretext task required in colorization \cite{Zhang2016Color} removes color as transformation and recovers the color information. Since both the methods map the images invariant to the transformation, they are considered as the alignment methods. Recently proposed vast majority of methods within this category are based on contrastive learning \cite{Oord2018CPC,He2019MoCo,Henaff2019CPC,Hjelm2019InfoMax,Chen2020SimCLR,Fedirici2020MIB,Misra2020PIRL,Tian2020CMC,Grill2020BYOL}, which generally avoids defining pretext tasks, and instead focuses on bringing representation of different views obtained by applying different transformations to the same image closer (`positive pairs'), and implicitly separating representations from different images (`negative pairs'). Contrastive methods often require comparing each example with many other examples to work well \cite{He2019MoCo}, which essentially indicates the importance of negative sampling. Contrastively, in this paper, we align different views or transformations by maximizing mutual information \cite{Hjelm2019InfoMax,Fedirici2020MIB} which is achieved by maximizing the lower bound of Jensen-Shenon divergence \cite{Hjelm2019InfoMax}. Additionally, to avoid shortcut or degenerating solution, we impose a bottlenecking regularization term inspired by information theory \cite{Tishby2015IB,Fedirici2020MIB}, which boosts our performance.

\myparagraph{Hybrid Methods:} 
Few other works have explored the combination of discriminative and alignment methods. Among them, Feng \etal \cite{Feng2019RotDecouple} show that a combination of the rotation prediction task \cite{Gidaris2018RotNet} and the alignment task through minimizing the representation distance with squared loss achieves encouraging results. Different to the existing hybrid methods, in this paper, we maximize mutual information as a way to align transformed images. Additionally, we use cropped transformations for the predictive tasks, which has also been proven to be competent in our experiment.

%TODO: Write in a better way
Our work is a hybrid method, where we define variant of transformations including crop, rotation, warp, blur, and discriminate only a subset of them. We ensure the transformations to contain mutually redundant information, which allow maximizing the mutual information. Addition to the discriminative task, we maximize their mutual information which is not same as merely minimizing representation distance; the presence of entropy within mutual information avoids degeneracy, as discussed in the following.

% Method
\begin{figure*}[!t]
\centering
\includegraphics[width=\textwidth]{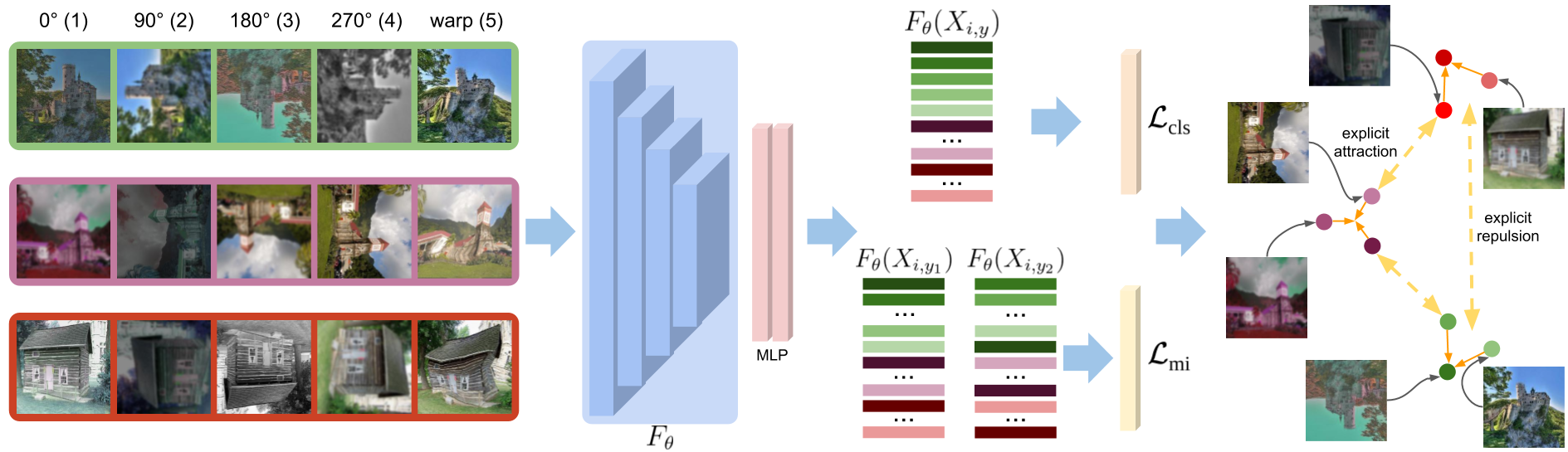}
%TODO: how to mention what is i and what is y? also reduce the following caption.
\caption{In our CODIAL model, we learn image representations $F_\theta$. In addition to auxiliary image transformations, we learn representations $F_\theta(X_{i,y})$ from primary transformations $X_{i,y}$ of images $X_i$. $F_\theta(X_{i,y})$ are directly utilized to train the model using the classification objective $\mathcal{L}_\text{cls}$ and they are paired to form tuples $(F_\theta(X_{i,y_1}), F_\theta(X_{i,y_2}))$ used for maximizing the mutual information criteria $\mathcal{L}_\text{mi}$. The joint discriminating and aligning criteria explicitly specifies which features should be repelled and which ones should be attracted.}
\vspace{-10pt}
\label{fig:arch_ssl}
\end{figure*}
% \zeynep{give a name to the method}

\section{Concurrent Discrimination and Alignment}
\label{sec:method}
%\TODO: Mutual information aligns views?? Yes, it does and we added one or two sentences in the intro and related works.
Our CODIAL framework learns unsupervised image representations by jointly solving two different pretext tasks: (1) recognizing the variations in global image statistics underwent by primary image transformations and (2) maximizing the mutual information of the image pairs formed with the transformed ones. \fig{fig:arch_ssl} depicts a comprehensive pipeline of our model considering images underwent with primary and auxiliary transformations $X_{i,y}$. 

%TODO: Improve the sentence that Zeynep commented
Our network model $F_\theta$ results in the transformed representations $F_\theta(X_{i,y})$ for each of the transformed images $X_{i,y}$ which are directly classified and trained with the objective in \eq{eqn:classification_loss}. The transformed representations $F_\theta(X_{i,y})$ are paired to form tuples $(F_\theta(X_{i,y_1}), F_\theta(X_{i,y_2}))$ to maximize the mutual information by minimizing the objective in \eq{eqn:mutual_information_loss}. The joint discrimination and alignment strategy only focuses on the generic object features, such as gradient, color, texture etc. which can be distinguished through transformations and are also common across views.

\subsection{Predicting Image Transformations}
\label{sec:image_transformation_prediction}
In addition to the primary transformations in \cite{Jenni2020GlobStat}, \ie rotation and warping, we use horizontal flipping, blurring, cropping and color jittering as auxiliary image transformations for creating deformed images so that the downstream representation becomes more robust to those distortions. 

Given a dataset of unlabeled images $S=\{X_i\}_{i=1}^N$ and a set of primary image transformations $T=\{t(X, y)\}_{y=1}^K$ for each image, we define $i$-th image with the $y$-th transformation as $X_{i,y}=t(X_i,y)$. A neural network model $F_\theta$ is trained to classify each transformed image to one of the $K$ primary transformations classes optimizing the following objective:
\begin{equation}
\mathcal{L}_\text{cls}=\min_\theta\frac{1}{KN}\sum_{i=1}^{N}\sum_{y=1}^{K}\ell_\text{cls}(F_\theta(X_{i,y});y)
\label{eqn:classification_loss}
\end{equation}
where $\ell_\text{cls}$ is the standard parametric cross-entropy loss for a multi-class classification problem. Particularly, we consider the images rotated with angles $0^\circ$, $90^\circ$, $180^\circ$, $270^\circ$ respectively as classes 1, 2, 3 and 4, and the warped image as class 5. Note that the auxiliary transformations are applied before the primary transformations. Particularly, random horizontal flipping is applied to the original image and the rest of the auxiliary transformations are applied before creating the individual transformed classes. Therefore note that the transformed images are not exactly the transformed version of the same image content. Intuitively, these variations in image transformations make the recognition task more difficult, which effectively makes the downstream representation more robust.

Predicting rotation is particularly effective for most natural images having objects in an up-front posture because any rotation of the image will result in an unusual object orientation. However, despite its simplicity and effectiveness, the assumption of rotation prediction task fails for the rotation-agnostic images \cite{Feng2019RotDecouple}. Similarly, an image warping is a smooth deformation of the image coordinates defined by $n$ pixel coordinates $\{(a_k, b_k)\}_{k=1,\ldots,n}$, which act as control points as represented in the following transformation
\begin{equation}
\begin{bmatrix}
a_k + \Delta a_k\\b_k + \Delta b_k \end{bmatrix} = \mathbf{T} \begin{bmatrix} a_k\\b_k\end{bmatrix}
\end{equation}
where $\Delta a_k$ and $\Delta b_k$ are the offsets and are uniformly sampled from a range $[-d, d]$, where $d=0.1 \times \text{dimension}(X_i)$. Image warping effectively changes the local image statistics only minimally and makes it difficult to distinguish a warped patch from a patch undergoing a change in perspective. Hence, the classifier needs to learn global image statistics to detect image warping. %\zeynep{it would look nicer if you could formalize some of these operations in an equation. Currently there is too much text.} %Simultaneously predicting angles of rotations and warping enables learning features from images covering a wide characteristics.

The main motivation behind the predictive classification task is to learn the generic features that distinguish those transformations and thus, effectively perform various downstream tasks. Image transformations, such as rotation \cite{Gidaris2018RotNet, Feng2019RotDecouple}, warping \cite{Jenni2018SSLArtifacts} change the global image statistics while the local visual information is kept intact. However, a network trained to solve a self-supervised task might accomplish it by using local statistics \cite{Doersch2015Context,Jenni2020GlobStat}. Such solutions are usually called \emph{shortcuts} and are a form of degenerate learning as they learn features with poor generalizations capabilities. We avoid those degenerate solutions by maximizing the mutual information between all possible pairs of transformed images as described next.

\subsection{Maximizing Mutual Information}
\label{sec:mutual_information}
Mutual information (MI) measuring relationship between random variables has often been used to learn the representation $F_\theta$. Let $X_{i,y_1}$, $X_{i,y_2}$ be a paired data sample from a joint probability distribution $P(X_{i,y_1}, X_{i,y_2})$, which is ensured as $X_{i,y_1}$ and $X_{i,y_2}$ are the two different transformed versions of the same image $X_i$. A representation $F_\theta$ can be learned by maximizing the mutual information between the encoded variables:
\begin{equation}
\max_\theta I(F_\theta(X_{i,y_1}); F_\theta(X_{i,y_2}))
\label{eqn:mutual_information}
\end{equation}
where $I$ denotes the mutual information and \eq{eqn:mutual_information} is equivalent to maximizing the predictability of $F_\theta(X_{i,y_1})$ from $F_\theta(X_{i,y_2})$, and vice versa. The aim of \eq{eqn:mutual_information} is to align the representations of paired samples. However, it is not the same as minimizing representation distance. In other words, the presence of entropy in the case of mutual information $I$ allows to avoid degenerate solutions.

We use the Jensen-Shannon mutual information estimator \cite{Hjelm2019InfoMax,Fedirici2020MIB}, which is a sample based differentiable mutual information lower bound. This procedure of maximizing MI needs introducing an additional parametric critic $C_\xi(F_\theta(X_{i,y_1}),F_\theta(X_{i,y_2}))$ which is jointly optimized with other parameters during the training procedure using re-parameterized samples from $F_\theta(X_{i,y_1})$ and $F_\theta(X_{i,y_2})$. Now, \eq{eqn:mutual_information} can trivially be solved by setting $F_\theta$ to the identity function because of the data processing inequality: 
\begin{equation}
    I(X_{i,y_1}; X_{i,y_2}) \ge I(F_\theta(X_{i,y_1}); F_\theta(X_{i,y_2})).
\end{equation} 
To alleviate the trivial solution, we impose a KL-divergence based regularizer rooted from the information bottleneck principle \cite{Tishby2015IB} which theoretically aligns the representations from $F_\theta(X_{i,y_1})$ to $F_\theta(X_{i,y_2})$, and vice versa, and discards transformation specific information as much as possible:
\begin{align}
R_\text{mib}(F_\theta(X_{i,y_1})&, F_\theta(X_{i,y_2})) = \frac{1}{2}D_\text{KL}(F_\theta(X_{i,y_1})||F_\theta(X_{i,y_2}))\nonumber\\
&+ \frac{1}{2}D_\text{KL}(F_\theta(X_{i,y_2})||F_\theta(X_{i,y_1}))
\label{eqn:mib_regularizer}
\end{align}
where $D_{KL}$ denotes the KL divergence for joint observation between two views. The above regularizer term when combined with the mutual information maximization objective, results in the following loss function:
\begin{align}
\mathcal{L}_\text{mi}=& -I(F_\theta(X_{i,y_1}); F_\theta(X_{i,y_2})) \nonumber \\
&+ \beta R_\text{mib}(F_\theta(X_{i,y_1}), F_\theta(X_{i,y_2}))
\label{eqn:mutual_information_loss}
\end{align}
where $\beta$ is the weight on the regularizer $R_\text{mib}$ and is increased during training from the initial value $10^{-6}$ to the final value $1.0$ with an exponential scheduling. Hence the final loss function that we aim to minimize is:
\begin{equation}
\mathcal{L} = \min_\theta \lambda_\text{cls}\mathcal{L}_\text{cls} + \lambda_\text{mi}\mathcal{L}_\text{mi}
\label{eqn:combined_loss}
\end{equation}
where $\lambda_\text{cls}$ and $\lambda_\text{mi}$ are respectively the weights on the classification and mutual information estimation criterion. For creating the pairs essential for estimating and maximizing mutual information, we consider all possible paired combinations of $K$ different transformations of the same image, which results in total $\nCr{K}{2}$ pairs. For efficiency, we uniformly sample a subset of size $k$ ($\le\nCr{K}{2}$) of such paired combinations. On the other hand, we empirically show that considering large number of such combinations increases the performance of our model. The critic network that estimates mutual information is adversarial and requires negative data points uniformly sampled from the transformed images different from the positive $\nCr{K}{2}$ pairs.

%TODO: Significance of Mutual Information
In addition to maximizing MI, we use information bottleneck regularizer \eq{eqn:mib_regularizer} \cite{Fedirici2020MIB} which discards superfluous information between the paired views. We carefully select auxiliary transformations, such as crop, blur, color jitter etc. for creating multiple views without affecting the label information. The transformed positive pairs are supposed to be mutually redundant. Intuitively, a view $X_{i,y_1}$ is redundant with respect to a second view $X_{i,y_2}$ whenever it is irrelevant for the label if $X_{i,y_2}$ is already observed, \ie in terms of mutual information $I(F_\theta(X_{i,y_1});X_{i,y_1}|X_{i,y_2})=0$. With the chain rule of MI:
\begin{align}
I(X_{i,y_1};F_\theta(X_{i,y_1})) &= I(X_{i,y_1};F_\theta(X_{i,y_1})|X_{i,y_2}) \nonumber \\ &+ I(X_{i,y_2};F_\theta(X_{i,y_1}))
\label{eqn:mi_chain_rule}
\end{align}
%
%Since our aim is to minimize transformation specific details without harming the label information, we simultaneously aim to minimize $I(X_{i,y_1};F_\theta(X_{i,y_1}))$ and maximize $I(X_{i,y_2};F_\theta(X_{i,y_1}))$. From \eq{eqn:mi_chain_rule}, 
where it is clear that $I(X_{i,y_1};F_\theta(X_{i,y_1}))$ can be reduced by minimizing $I(X_{i,y_1};F_\theta(X_{i,y_1})|X_{i,y_2})$ as the information $F_\theta(X_{i,y_1})$ contains is unique to $X_{i,y_1}$ and is not predictable by observing $X_{i,y_2}$. However, since we assume that mutual redundancy between $X_{i,y_1}$ and $X_{i,y_2}$, $I(X_{i,y_1};F_\theta(X_{i,y_1})|X_{i,y_2})$ is irrelevant and can be safely discarded, our first loss:
\begin{align}
&\mathcal{L}_1(\theta,\lambda_1) = \nonumber \\ &I(F_\theta(X_{i,y_1});X_{i,y_1}|X_{i,y_2}) - \lambda_1 I(X_{i,y_2};F_\theta(X_{i,y_1}))
\end{align}
where $\lambda_1$ is the Lagrangian multiplier introduced by the constrained optimization. Our second loss:
\begin{align}
&\mathcal{L}_2(\theta,\lambda_2) = \nonumber \\ &I(F_\theta(X_{i,y_2});X_{i,y_2}|X_{i,y_1}) - \lambda_2 I(X_{i,y_1};F_\theta(X_{i,y_2}))
\end{align}
By re-parametrizing the Lagrangian multipliers, the average of the two loss functions $\mathcal{L}_1$ and $\mathcal{L}_2$ can be proven to be upper bounded as in \eq{eqn:mutual_information_loss} \cite{Fedirici2020MIB}. This allows learning invariances directly from the augmented data, rather than requiring them to be built into the model architecture.

%Zeynep: this paragraph is repeating the last paragraph of the last sub section
%TODO: Improve non-degenerate Solutions
%To discriminate several image transformations, it is important to make sure that the trained network cannot exploit (local) artifacts introduced by the transformations to solve the task. For example, for face datasets, only determining the orientation of a subset of local patches (eye, lips, nose etc.) could decide the orientation of the whole image, but this may result in a representation that only focuses local statistics. Moreover, for some images determining the correct orientation is an unambiguous task and hence this may not improve robustness. Therefore, maximizing mutual information between the transformed pairs of the same image removes the risk of such degenerate solutions. %based on local features and unambiguous prediction task.

% Experimental Validation
{
\setlength{\tabcolsep}{4pt}
\renewcommand{\arraystretch}{1.1}
\begin{table*}[!ht]
\centering
\resizebox{\linewidth}{!}{
\begin{tabular}{l |ccccc | ccccc  | c | c | c}
& \multicolumn{5}{c|}{ \textbf{STL-10}}  & \multicolumn{5}{c|}{ \textbf{CelebA}} & \multirow{2}{*}{\textbf{CIFAR-10}} & \multirow{2}{*}{\textbf{CIFAR-100}} & \textbf{Tiny-}  \\
\textbf{Method} & \texttt{c1} & \texttt{c2} & \texttt{c3} & \texttt{c4} & \texttt{c5} & \texttt{c1} & \texttt{c2} & \texttt{c3} & \texttt{c4} & \texttt{c5} & & & \textbf{ImageNet} \\
\hline
RotNet \cite{Gidaris2018RotNet} & 58.2 & 67.3 & 69.3 & 69.9 & 70.1 & 70.3 & 70.9 & 67.8 & 65.6 & 62.1
& 62.1 & 33.2 & 23.7 \\
Decouple \cite{Feng2019RotDecouple} & 59.0 & 68.6 & 70.8 & 72.5 & 74.6 & 71.4 & 73.8 & 72.7 & 72.4 & 72.2 & 65.9 & 37.7 & 28.5 \\
GlobStat \cite{Jenni2020GlobStat} & 59.2 & 69.7 & 71.9 & 73.1 & 73.7 & 71.8 & 74.0 & 73.5 & 72.5 & 69.2 & 68.1 & 39.2 & 31.2 \\
CODIAL (ours) & \textbf{60.5} & \textbf{71.5} & \textbf{74.3} & \textbf{75.3} & \textbf{75.4} & \textbf{84.8} & \textbf{86.1} & \textbf{84.9} & \textbf{83.8} & \textbf{82.7}
& \textbf{69.9} & \textbf{43.7} & \textbf{33.6} \\
%\bottomrule
\end{tabular}
}
\caption{Comparing with the state-of-the-art on STL-10, CelebA, CIFAR-10, CIFAR-100 and TinyImageNet datasets. For STL-10 and CelebA, we report results obtained by different convolutional layers of the backbone AlexNet (\texttt{c1-5}). The results report test set performance of the linear classifiers on the remaining datasets trained on the frozen layers of AlexNet models pre-trained with the pretext tasks.}
% \zeynep{there needs to be space between the text and the table. please do not alter the template.} \anjan{sorry, I shall not change anymore.}
\vspace{-10pt}
\label{tab:stl10_results}
\end{table*}
}

\section{Experiments}
\label{sec:expt}
In this section, we perform an extensive experimental evaluation of our model on several unsupervised feature learning benchmarks and ablate our model components.

\myparagraph{Datasets:} We evaluate our model on 9 benchmark datasets: STL-10 \cite{Coates2011STL10},
CIFAR-10 \cite{Krizhevsky2009CIFAR},
CIFAR-100 \cite{Krizhevsky2009CIFAR},
TinyImageNet \cite{Li2015TinyImageNet},
ImageNet-1K \cite{Russakovsky2014ImageNet},
Places 205 \cite{Zhou2014Places},
CelebA \cite{Liu2015CelebA},
PASCAL VOC 2007 \cite{Everingham2014PASCAL},
PASCAL VOC 2012 \cite{Everingham2014PASCAL}. Among these, STL-10 (10 classes, 100K unlabeled and 5K labeled images for training, 8K for test) and CIFAR-10 (10 classes, 60K images) are small scale, CIFAR-100 (100 classes, 60K images) and TinyImageNet (200 classes, 110K images) are medium scale, and ImageNet-1K (1K classes, 1.2M images) and Places 205 (205 classes, 2.4M images) are large scale classification datasets with a single label per image. CelebA (40 classes, 182K images) is a facial attribute dataset while PASCAL VOC 2007 (20 classes, 5K training, 5K test images) and PASCAL VOC 2012 (21 classes, 10K training, 10K test images) are image classification, object detection and semantic segmentation datasets where there are multiple labels per image. On STL-10, CIFAR-10, CIFAR-100, TinyImageNet, ImageNet-1K and Places 205, the evaluation metric is top-1 accuracy. On CelebA and PASCAL VOC 2007, the evaluation metric is mAP (mean average precision) and on PASCAL VOC 2012, it is mIoU (mean intersection over union).
% \zeynep{correct?} \anjan{yes, it is correct, thank you :)}

\myparagraph{Implementation Details:}
For a fair comparison with prior works, we implement our CNN model $F_\theta$ as the standard AlexNet architecture \cite{Krizhevsky2012AlexNet}. Following prior works \cite{Gidaris2018RotNet,Feng2019RotDecouple,Jenni2020GlobStat}, we remove the local response normalization layers and add batch normalization to all layers except for the final one. No other modifications to the original architecture are made. For experiments on lower resolution images (\eg, STL-10), we remove the max-pooling layer after \texttt{conv5} and use the default padding setting throughout the network. The auxiliary data augmentation strategies (random cropping, horizontal flipping, color jittering and blurring) are used and validated through experiments in the following sections. The size of the patch boundary is set to 2 pixels in experiments on STL-10 and CelebA. On ImageNet-1K, we use a 4 pixel boundary. %We simply set the parameters $\lambda_\text{cls}$, $\lambda_\text{mi}$ to 1, which is also corroborated through experiments in the following section.

\subsection{Comparing with the State-of-the-Art}

%In this section, we compare our obtained results on five datasets with three recently proposed methods including discriminative as well as hybrid one. Among them, RotNet \cite{Gidaris2018RotNet} discriminating rotation angles and GlobStat \cite{Jenni2020GlobStat} distinguishing rotation angles, warping and limited context inpainting are considered as discriminating methods. Feat Decouple \cite{Feng2019RotDecouple} jointly discriminates rotation angles and align similar features by minimizing representation distance is an instance of hybrid method.

\myparagraph{Methods:} To evaluate the effectiveness of our hybrid CODIAL model that follows self-supervised learning protocol, we compare it with three state-of-the-art methods on five datasets. RotNet \cite{Gidaris2018RotNet} discriminating rotation angles and GlobStat \cite{Jenni2020GlobStat} distinguishing rotation angles, warping and limited context inpainting are considered as discriminating methods. Decouple \cite{Feng2019RotDecouple} jointly discriminates rotation angles and aligns similar features by minimizing representation distance is an instance of hybrid method.

%Anjan: having the above paragraph on methods, we can discard the following lines: We compare our obtained results on five datasets with three recent methods proposed by Gidaris \etal \cite{Gidaris2018RotNet}, Feng \etal \cite{Feng2019RotDecouple} and Jenni \etal \cite{Jenni2020GlobStat}, following the same experimental protocol for all the three methods. Note that these prior works include discriminative methods \cite{Gidaris2018RotNet,Jenni2020GlobStat} as well as hybrid methods \cite{Feng2019RotDecouple}.}

\myparagraph{Results:} We observe in \tab{tab:stl10_results} that our method steadily improves over all the prior methods' reported results on these datasets. Specifically, the consistency over all the AlexNet layers on both the datasets shows the robustness of our model to learn low level image features to high level semantic features. On CelebA, our model particularly outperforms the prior works by a large margin as a mutual benefit of our proposed discriminative and aligning objectives. We also observe that on CelebA dataset, similar to other prior works, our model performs better in lower layers as compared to the higher layers, as lower level geometric features are more important for recognizing different facial attributes than the high level semantic features of the higher layers. On the challenging STL-10, our method respectively outperforms recently proposed Decouple and GlobStat methods by an average margins of 1.9\% and 2.1\%, and on CelebA, it respectively obtains a margin of 12.3\% and 12.0\%. This indicates the precedence of our hybrid method that concurrently discriminates and aligns features, and the mutual information based alignment strategy.

Our CODIAL surpasses the self-supervised benchmarks on CIFAR-10, CIFAR-100 and TinyImageNet datasets with a good margin, which further emphasizes the benefit of our hybrid learning approach. Particularly, we surpass GlobStat method on CIFAR-10, CIFAR-100 and TinyImageNet datasets respectively by the margins of 1.8\%, 4.5\% and 2.4\% which shows the effectiveness of our method. We exceed Decouple respectively by the margins of 4.0\%, 6.0\% and 5.1\% on the same datasets, as an advantage of our MI based alignment strategy, since Decouple aligns features by minimizing representation distance.
% \massi{(M:Here we might highlight what the baselines do (or at least their category) to later explicitly state something like: "we outperform \cite{Gidaris2018RotNet} because its method does not consider alignment" etc.)}

% \newpage

{
\setlength{\tabcolsep}{2.5pt}
\renewcommand{\arraystretch}{1.2}
\begin{table}[t]
\centering
 \resizebox{\linewidth}{!}{
\begin{tabular}{l|ccccc |ccccc}
%\toprule
  & \multicolumn{5}{c|}{ \textbf{STL-10}}  & \multicolumn{5}{c}{ \textbf{CelebA}}  \\
  \hline
\textbf{Transf.} & \texttt{c1} & \texttt{c2} & \texttt{c3} & \texttt{c4} & \texttt{c5} & \texttt{c1} & \texttt{c2} & \texttt{c3} & \texttt{c4} & \texttt{c5}\\
\hline
cr &  59.9 & 69.8 & 73.4 & 73.9 & 74.3 & 84.7 & 85.7 & 84.0 & 81.8 & 79.5 \\
%\midrule
cr + bl & 59.5 & 69.5 & 71.9 & 72.8 & 72.9 & 84.6 & 85.8 & 84.1 & 81.9 & 79.8 \\
cr + co & \textbf{60.5} & \textbf{71.5} & \textbf{74.3} & \textbf{75.3} & \textbf{75.4} & 84.8 & \textbf{86.1} & \textbf{84.9} & \textbf{83.8} & \textbf{82.7} \\
%\midrule
cr + bl + co & 59.6 & 71.0	& 73.2 & 74.8 & 75.1 & \textbf{84.9} & 85.9 & \textbf{84.9} & 83.7 & 82.5 \\
\hline
\textbf{Loss} & \texttt{c1} & \texttt{c2} & \texttt{c3} & \texttt{c4} & \texttt{c5} & \texttt{c1} & \texttt{c2} & \texttt{c3} & \texttt{c4} & \texttt{c5}\\
\hline
Random & 48.4 & 53.3 & 51.1 & 48.7 & 47.9 & 68.9 & 70.1 & 66.7 & 65.3 & 63.2\\
%\midrule
$\lambda_{\text{cls},\text{mi}}$ = 1,0 & 57.8 & 68.7 & 69.2 & 68.1 & 66.4 & 74.5 & 75.5 & 74.0 & 72.6 & 71.5\\
$\lambda_{\text{cls},\text{mi}}$ = 0,1 & 58.2 & 67.0 & 68.9 & 67.9 & 66.5 & 71.2 & 73.7 & 72.2 & 69.4 & 67.7\\
$\lambda_{\text{cls},\text{mi}}$ = 1,1 & \textbf{60.5} & \textbf{71.5} & \textbf{74.3} & \textbf{75.3} & \textbf{75.4} & \textbf{84.8} & \textbf{86.1} & \textbf{84.9} & \textbf{83.8} & \textbf{82.7}
\end{tabular}
 }
\caption{Ablating different transformations (top): We report test set performance of our pre-text model trained on different transformations. Ablating the loss weights (bottom): We report the test set performance of our final model by varying the loss weights. (STL-10 and CelebA with CNN Backbone AlexNet with conv layers (\texttt{c1-5}), cr: crop, bl: blur, co: color transformations).}
\label{tab:stl10_results_transforms}
\vspace{-12pt}
\end{table}
}

\subsection{Ablating Model Components}
\label{sec:preliminary_analysis}
We perform our ablation study on STL-10 and CelebA datasets and analyze different design choices. For all the experiments in this section, we first pre-train our model for solving the self-supervised pretext tasks then we employ linear classifiers on top of the frozen convolutional features to classify the images from the respective datasets. %, \ie, linear classification of STL-10 images is done by considering the pre-task model trained on the STL-10 dataset itself.

\myparagraph{Auxiliary Image Transformations:} We consider several auxiliary image transformations or augmentations, such as random horizontal flipping, random cropping, color jittering and blurring. Among them, random cropping is done by selecting a random patch of the image with an area uniformly sampled between 8\% and 100\% of that of the original image, and an aspect ratio logarithmically sampled between 3/4 and 4/3. This patch is then resized to a squared image of size dependent on the datasets. In case of the color jittering transform, we shift brightness, contrast, saturation, and hue of each pixel in the image by a uniformly sampled offset from the range $[0.5, 1.5]$. The order in which these shifts are performed is randomly selected for each image. Gaussian blurring is implemented with a Gaussian kernel of size 10\% of that of the image and a standard deviation uniformly sampled over $[0.1, 2.0]$.
In this experiment, we systematically analyze the contributions of the above mentioned image transformations on our model. For doing so, we pre-train our model with the above augmentations on the STL-10 and CelebA datasets and report the performance of linear classifiers trained on top of the frozen AlexNet layers.

From \tab{tab:stl10_results_transforms} (top) we observe that combining color jittering with other transformations improves the results across all the AlexNet layers on both the datasets, however, the difference is more in case of STL-10 as it contains natural images with diverse colors compared to CelebA dataset which mainly contains face images where only skin color has a prevalence. As a result, color jittering produces better outcome on STL-10 compared to CelebA dataset. We also observe that on STL-10, the difference of accuracy with \texttt{conv1}, \texttt{conv2}, and \texttt{conv3} features are respectively 1.0\%, 2.0\% and 2.4\%, while on CelebA dataset, the difference of mAP values are 0.1\%, 0.4\% and 0.9\%. For CelebA dataset, we speculate that since our self-supervised features are intend to learn high level semantic features, it does not waver much the geometric features.

% although for the lower layers (\eg, \texttt{conv1}, \texttt{conv2}, \texttt{conv3}) the difference in performance is not very significant for different image augmentations. However, for higher layers (\eg, \texttt{conv4}, \texttt{conv5}) the difference is quite notable, and we discover that combining color jittering with other transformations improves the results for \texttt{conv4} and \texttt{conv5} layers.

% \anjan{explanation/conjecture needed}.

% \massi{(M: I do not know if we can comment on this with e.g., why jittering+blurring and cropping is the best. If I am understanding correctly, another thing is that in Tab.\ref{tab:stl10_results_transforms} and \ref{tab:celeba_results_transforms} the results show that actually ignoring blurring works as well or even better then using the three of them altogether. In case we used the combination of the three we could justify why.)} \anjan{Yes, that's right and I corrected that.}

\begin{figure}[!t]
\resizebox{\columnwidth}{!}{
\begin{tabular}{cc}
\includegraphics[]{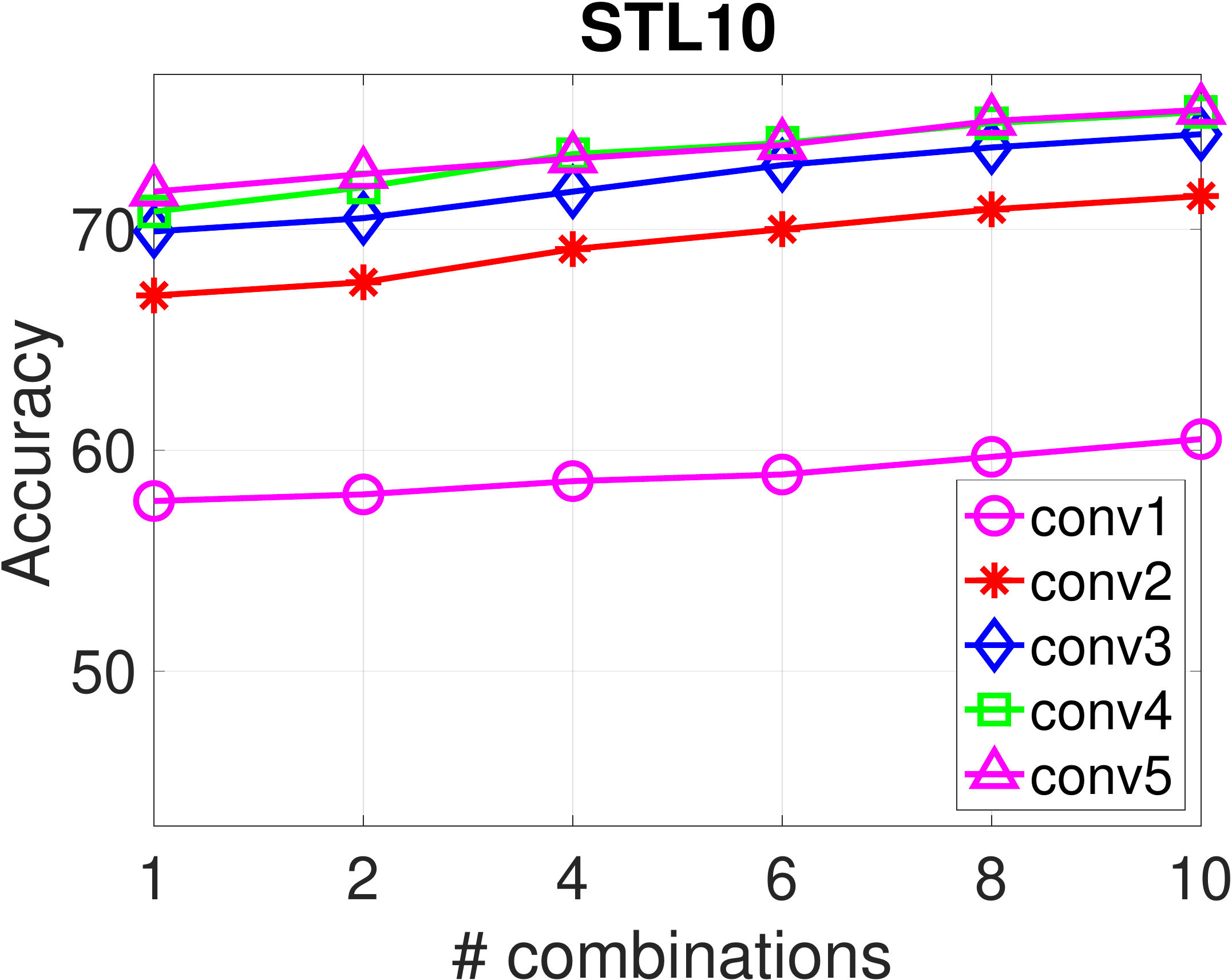} & \includegraphics[]{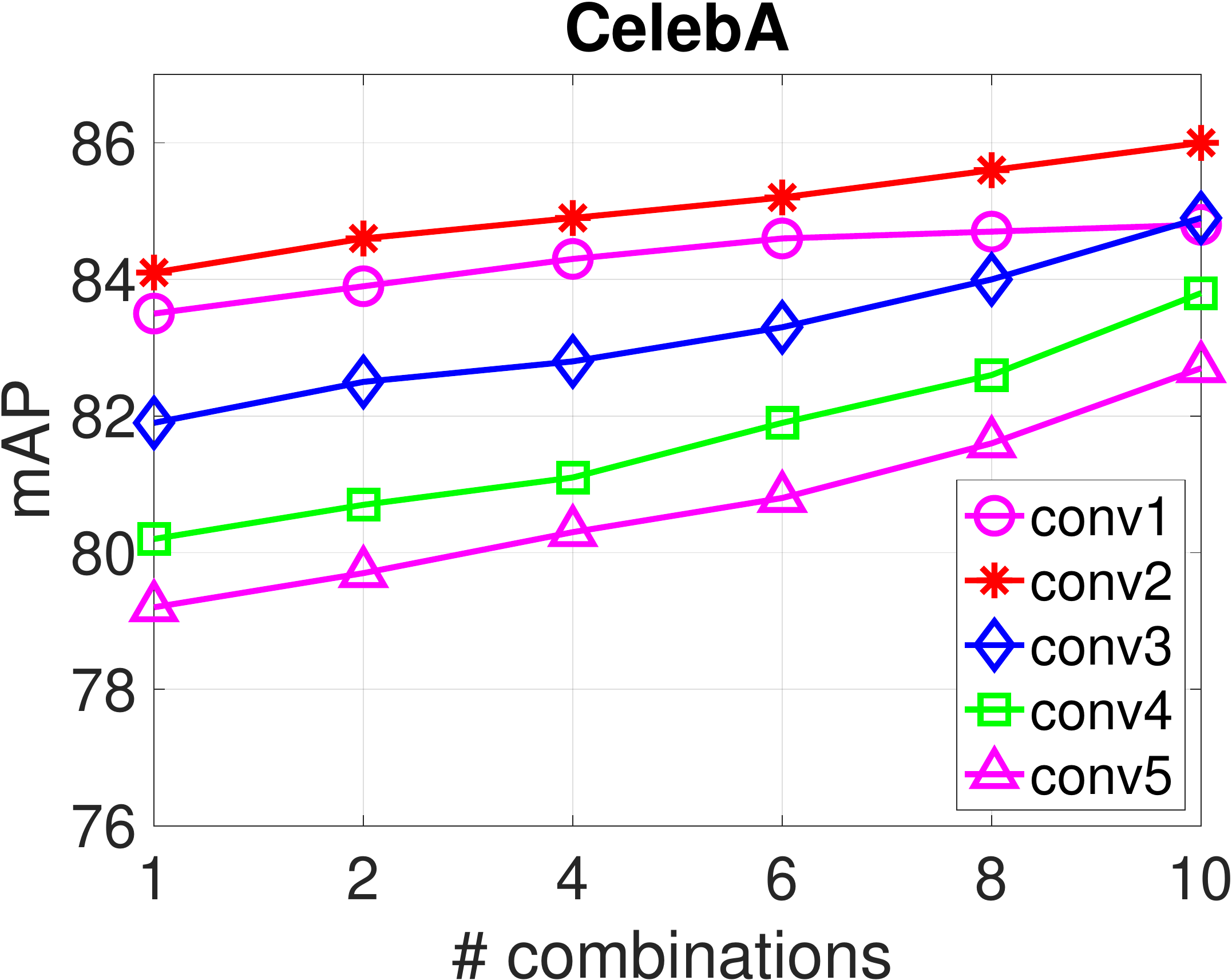}
\end{tabular}}
\caption{Plots showing the performance of five convolutional layers of our AlexNet model trained with increasing number of paired transforms, exhibit that a large number of transformed pairs is important for better performance.}
\label{fig:results_multiple_transforms}
\vspace{-12pt}
\end{figure}

{
\setlength{\tabcolsep}{6pt}
\renewcommand{\arraystretch}{1.1}
\begin{table*}[t]
\centering
\resizebox{0.9\linewidth}{!}{
\begin{tabular}{l| ccccc| ccccc|| ccc}
%\toprule
& \multicolumn{5}{c|}{ \textbf{ImageNet-1K}}  & \multicolumn{5}{c||}{ \textbf{Places 205}} & \multicolumn{3}{c}{ \textbf{Pascal VOC}}  \\
\textbf{Model / Layer} & \texttt{c1} & \texttt{c2} & \texttt{c3} & \texttt{c4} & \texttt{c5} & \texttt{c1} & \texttt{c2} & \texttt{c3} & \texttt{c4} & \texttt{c5} & cls. & det. & seg.\\
\hline
Places Labels & \multicolumn{5}{c|}{Not applicable} & 22.1 & 35.1 & 40.2 & 43.3 & 44.6 & \multicolumn{3}{c}{Not applicable}\\
ImageNet Labels & 19.3 & 36.3 & 44.2 & 48.3 & 50.5 & 22.7 & 34.8 & 38.4 & 39.4 & 38.7 & 79.9 & 59.1 & 48.0\\
Random & 11.6 & 17.1 & 16.9 & 16.3 & 14.1 & 15.7 & 20.3 & 19.8 & 19.1 & 17.5 & 53.3 & 43.4 & 19.8\\
\hline
DeepCluster \cite{Caron2018DeepClustering} & 12.9 & 29.2 & 38.2 & 39.8 & 36.1 & 18.6 & 30.8 & 37.0 & 37.5 & 33.1 & 73.7 & 55.4 & 45.1\\
RotNet \cite{Gidaris2018RotNet} & 18.8 & 31.7 & 38.7 & 38.2 & 36.5 & 21.5 & 31.0 & 35.1 & 34.6 & 33.7 & 73.0 & 54.4 & 39.1\\
KnowTrans \cite{Noroozi2018SSLKT} & 19.2 & 32.0 & 37.3 & 37.1 & 34.6 & 19.2 & 32.0 & 37.3 & 37.1 & 34.6 & 72.5 & 56.5 & 42.6\\
Decouple \cite{Feng2019RotDecouple} & 19.3 & 33.3 & 40.8  & 41.8 & \textbf{44.3} & 22.9 & 32.4 & 36.6 & 37.3 & \textbf{38.6} & 74.3 & 57.5 & 45.3\\
SpotArtifacts \cite{Jenni2018SSLArtifacts} & 19.5 & 33.3 & 37.9 & 38.9 & 34.9 & 23.3 & 34.3 & 36.9 & 37.3 & 34.4 & 74.3 & 57.5 & 45.3\\
GlobStat \cite{Jenni2020GlobStat} & 20.8 & 34.5 & 40.2 & 43.1 & 41.4 & 24.1 & 33.3 & 37.9 & 39.5 & 37.7 & 74.5 & 56.8 & 44.4\\
CMC \cite{Tian2020CMC} & 18.3 & 33.7 & 38.3 & 40.5 & 42.8 & - & - & - & - & - & 73.8 & 56.6 & 44.8 \\
% \rowcolor{babyblueeyes}
CODIAL (ours) & \textbf{22.1} & \textbf{36.2} & \textbf{41.8} & \textbf{44.3} & 42.2 & \textbf{25.5} & \textbf{34.6} & \textbf{38.9} & \textbf{41.2} & 38.4 & \textbf{75.2} & \textbf{58.3} & \textbf{45.5} \\
%\bottomrule
\end{tabular}
}
\caption{Validation set accuracy (\%) on ImageNet-1K and Places 205 datasets (left) with linear classifiers trained on frozen convolutional layers and transfer learning results for classification, detection (mAP) and segmentation (mIoU) on PASCAL VOC (right) compared to state-of-the-art feature learning methods (\texttt{c1-5}: convolutional layers 1-5 of AlexNet, cls: classification, det: detection, seg: segmentation).}
\label{tab:imagenet_results}
\vspace{-12pt}
\end{table*}
}

\myparagraph{Loss function:} We compare the impact of two parts of our proposed loss function (\eq{eqn:combined_loss}). For that, we pre-train our model by selecting the weights of $\lambda_\text{cls}$ and $\lambda_\text{mi}$ from $\{0,1\}$ in \eq{eqn:combined_loss} on the STL-10 and CelebA datasets and train a linear classifier on top of different frozen layers of AlexNet.

As shown in \tab{tab:stl10_results_transforms} (bottom), considering the discriminative part together with the mutual information maximization substantially boosts the performance of the method. Specifically, on STL-10 dataset, we obtain 2.7\%, 2.8\%, 5.1\%, 7.2\% and 9.0\% accuracy boost and on CelebA dataset we obtain 10.3\%, 10.6\%, 10.9\%, 11.2\%, 11.2\% mAP raise respectively on the \texttt{conv1}, \texttt{conv2}, \texttt{conv3}, \texttt{conv4} and \texttt{conv5} features, which is quite significant. This can be seen as a benefit of explicitly specifying which features should be close to or far from each other.
% \anjan{explanation/conjecture needed}
% \massi{(M: It might be helpful to stress the importance/mutual benefit of the two parts already here. At the end this is the core message of the work.}

\myparagraph{Multiple Image Transformations:} By design, our model can exploit more than one pair of transformations of the same image. We design this experiment to verify whether considering more than one image pair has any benefit on the performance of the model. For that, we choose the same STL-10 and CelebA datasets, and pre-train our model with a subset of pairs uniformly selected from all the 10 ($\nCr{K}{2}$ and $K=5$ in our case) transformed pairs. Once trained, the different layers of our AlexNet model are evaluated on the same downstream task on the same STL-10 and CelebA datasets. \fig{fig:results_multiple_transforms} shows the performance of different layers of AlexNet for a variant number of paired transformations. The plots shown in the figure exhibits climbing trend for all the layers as the number of paired transformations increases and for all the layers the best performances are obtained when all possible transformed pairs are considered. This indicates that a large number of transformed pairs is important for effectively maximizing mutual information and our model design constructively supports that necessity.

\subsection{Unsupervised Feature Learning}
\label{sec:unsupervised_benchmarks}
In addition to RotNet \cite{Gidaris2018RotNet}, Decouple \cite{Feng2019RotDecouple} and GlobStat \cite{Jenni2020GlobStat}, we compare our method with four more recently proposed methods. Among them, DeepCluster \cite{Caron2018DeepClustering} iteratively clusters the features with kmeans algorithm, and uses the cluster labels as supervision to update the weights of the network; KnowTrans \cite{Noroozi2018SSLKT} uses clustering to boost the self-supervised knowledge transfer; SpotArtifacts \cite{Jenni2018SSLArtifacts} learns self-supervised knowledge by spotting synthetic artifacts in images; CMC \cite{Tian2020CMC} learns self-supervised knowledge by contrasting multiple view of the same data. We pre-train our model for 100 epochs on ImageNet-1K \cite{Russakovsky2014ImageNet}, where the images are cropped to $224 \times 224$. The pre-training was done with batch size of 256 and on 2 Titan RTX GPUs.
% \massi{(M: Here I would put a brief descriptions of the sota methods)}

\myparagraph{Linear Classification on ImageNet-1K and Places 205:} Following \cite{Zhang2016Color,Feng2019RotDecouple,Jenni2020GlobStat}, we train linear classifiers on top of the frozen features extracted by different convolutional layers measuring the task specific power of the learned representations, specifically the discriminative power over object class. %As usual, we perform this study on both the ImageNet-1K \cite{Russakovsky2014ImageNet} as well as the Places 205 dataset \cite{Zhou2014Places}.
For the experiments on the ImageNet-1K and Places 205 datasets in \tab{tab:imagenet_results} (left), all the approaches use AlexNet as the backbone network and all the methods except `ImageNet Labels', `Places Labels' and `Random' are pre-trained on ImageNet-1K without labels. `ImageNet Labels' and `Places Labels' are supervised benchmarks and are respectively trained with ImageNet-1K and Places 205 labels. All the weights of the feature extractor network are frozen and feature maps are spatially resized (with adaptive pooling) so as to have the feature dimension around 9,000.
%Anjan: prior works (\eg Jenni \etal, Feng \etal etc.) enlisted many more previous methods, many of which are quite old works from 2015, 2016. We can put them, if we have space left.

Our learned features appear to be very robust and achieve state-of-the-art performance with all the layers from \texttt{conv1} to \texttt{conv4} on ImageNet-1K, particularly on \texttt{conv4} features, we obtain 1.2\% improvement that GlobStat, which is quite remarkable. Our result on \texttt{conv5} also surpasses the recent benchmark GlobStat \cite{Jenni2020GlobStat} by 0.8\% and is comparable with the best result obtained by Decouple \cite{Feng2019RotDecouple}. Considering the fact that the lower layers of a network usually capture low-level information like edges or contours in images and the higher layers extract abstract semantic information, our model shows diverse capacity in capturing low-level as well as high-level abstract features which can be considered as the benefit of our hybrid model. We also observe the same on the Places 205 dataset and achieve the state-of-the-art results with all layers from \texttt{conv1} to \texttt{conv4}. Our result on \texttt{conv4} in particular is the overall best among all the methods except the \texttt{conv4} and \texttt{conv5} layers of `Places Labels' which is a fully supervised model. Note, that our method surpasses the performance of an AlexNet trained on ImageNet-1K using supervision on the \texttt{conv1}, \texttt{conv3} and \texttt{conv4} layers respectively by 2.8\%, 0.5\%, 1.8\%.

\begin{figure*}[!ht]
\begin{center}
\resizebox{\textwidth}{!}{
\begin{tabular}{@{}cc@{}c@{}c@{}c@{}c@{}c@{}c@{}cc@{}c@{}c@{}c@{}c@{}c@{}c@{}cc@{}c@{}c@{}c@{}c@{}c@{}c@{}c}
\LARGE{Query} & & \multicolumn{7}{c}{\LARGE{Supervised / ImageNet Labels \cite{Krizhevsky2012AlexNet}}} & & \multicolumn{7}{c}{\LARGE{GlobStat \cite{Jenni2020GlobStat}}} & & \multicolumn{7}{c}{\LARGE{CODIAL (Ours)}}\\
\includegraphics[width=1.5cm, height=1.5cm]{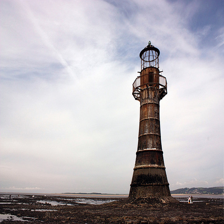} & & \includegraphics[width=1.5cm, height=1.5cm]{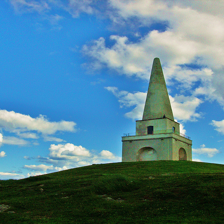} & \includegraphics[width=1.5cm, height=1.5cm]{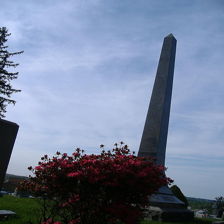} & \includegraphics[width=1.5cm, height=1.5cm]{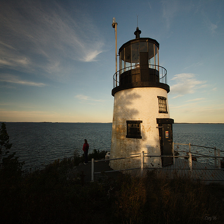} & \includegraphics[width=1.5cm, height=1.5cm]{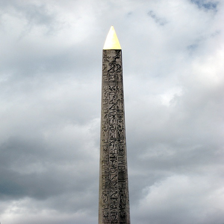} & \includegraphics[width=1.5cm, height=1.5cm]{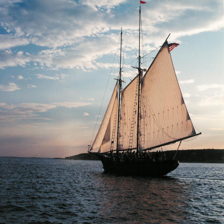} & \includegraphics[width=1.5cm, height=1.5cm]{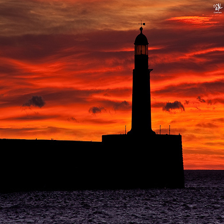} & \includegraphics[width=1.5cm, height=1.5cm]{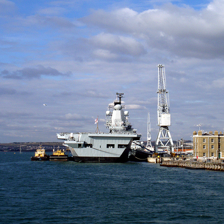} & & \includegraphics[width=1.5cm, height=1.5cm]{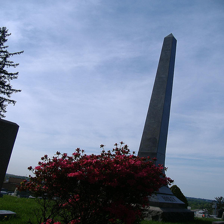} & \includegraphics[width=1.5cm, height=1.5cm]{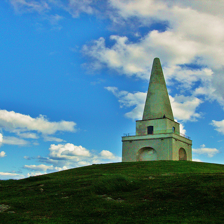} & \includegraphics[width=1.5cm, height=1.5cm]{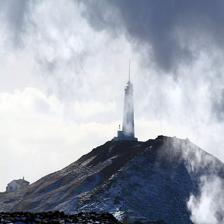} & \includegraphics[width=1.5cm, height=1.5cm]{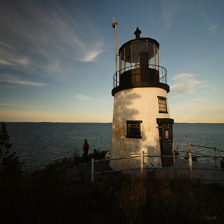} & \includegraphics[width=1.5cm, height=1.5cm]{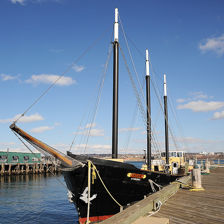} & \includegraphics[width=1.5cm, height=1.5cm]{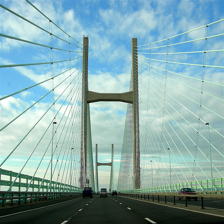} & \includegraphics[width=1.5cm, height=1.5cm]{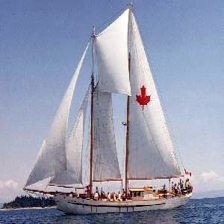} & & \includegraphics[width=1.5cm, height=1.5cm]{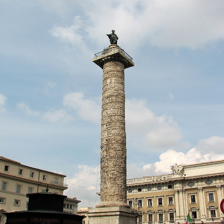} & \includegraphics[width=1.5cm, height=1.5cm]{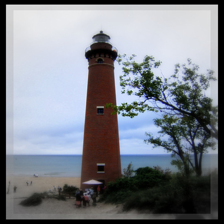} & \includegraphics[width=1.5cm, height=1.5cm]{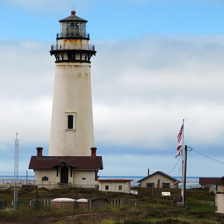} & \includegraphics[width=1.5cm, height=1.5cm]{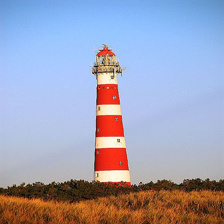} & \includegraphics[width=1.5cm, height=1.5cm]{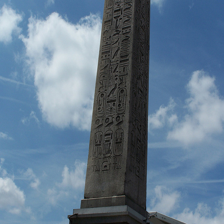} & \includegraphics[width=1.5cm, height=1.5cm]{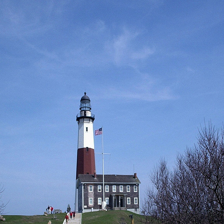} & \includegraphics[width=1.5cm, height=1.5cm]{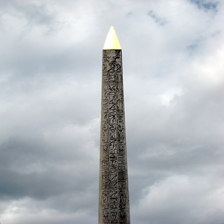}\\
& & \cmark & \cmark & \cmark & \cmark & \xmark & \cmark & \xmark & & \cmark & \cmark & \cmark & \cmark & \xmark & \xmark & \xmark & & \cmark & \cmark & \cmark & \cmark & \cmark & \cmark & \cmark\\
\includegraphics[width=1.5cm, height=1.5cm]{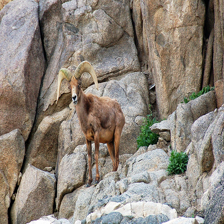} & & \includegraphics[width=1.5cm, height=1.5cm]{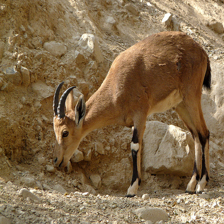} & \includegraphics[width=1.5cm, height=1.5cm]{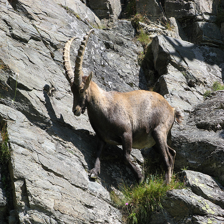} & \includegraphics[width=1.5cm, height=1.5cm]{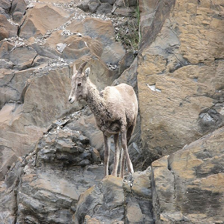} & \includegraphics[width=1.5cm, height=1.5cm]{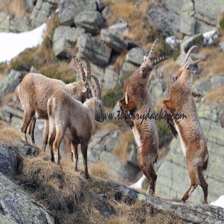} & \includegraphics[width=1.5cm, height=1.5cm]{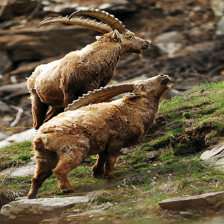} & \includegraphics[width=1.5cm, height=1.5cm]{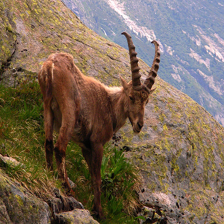} & \includegraphics[width=1.5cm, height=1.5cm]{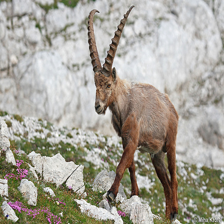} & & \includegraphics[width=1.5cm, height=1.5cm]{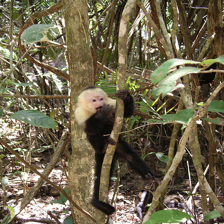} & \includegraphics[width=1.5cm, height=1.5cm]{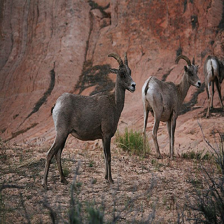} & \includegraphics[width=1.5cm, height=1.5cm]{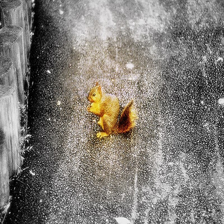} & \includegraphics[width=1.5cm, height=1.5cm]{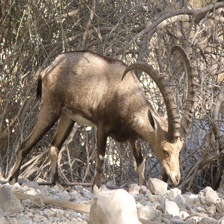} & \includegraphics[width=1.5cm, height=1.5cm]{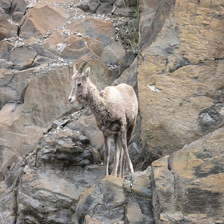} & \includegraphics[width=1.5cm, height=1.5cm]{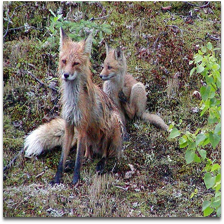} & \includegraphics[width=1.5cm, height=1.5cm]{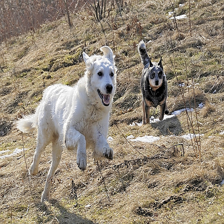} & & \includegraphics[width=1.5cm, height=1.5cm]{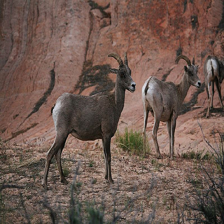} & \includegraphics[width=1.5cm, height=1.5cm]{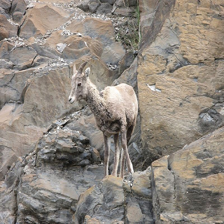} & \includegraphics[width=1.5cm, height=1.5cm]{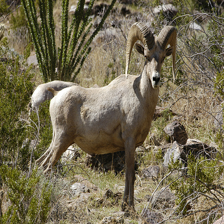} & \includegraphics[width=1.5cm, height=1.5cm]{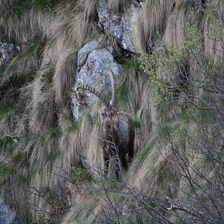} & \includegraphics[width=1.5cm, height=1.5cm]{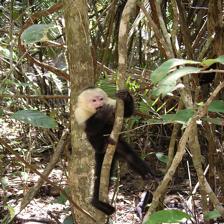} & \includegraphics[width=1.5cm, height=1.5cm]{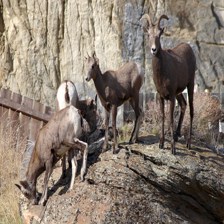} & \includegraphics[width=1.5cm, height=1.5cm]{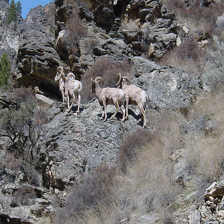}\\
& & \cmark & \cmark & \cmark & \cmark & \cmark & \cmark & \cmark & & \cmark & \cmark & \xmark & \cmark & \cmark & \xmark & \xmark & & \cmark & \cmark & \cmark & \cmark & \cmark & \cmark & \cmark\\
% \includegraphics[width=1.5cm, height=1.5cm]{03/query} & & \includegraphics[width=1.5cm, height=1.5cm]{03/supervised/1_1} & \includegraphics[width=1.5cm, height=1.5cm]{03/supervised/2_1} & \includegraphics[width=1.5cm, height=1.5cm]{03/supervised/3_1} & \includegraphics[width=1.5cm, height=1.5cm]{03/supervised/4_1} & \includegraphics[width=1.5cm, height=1.5cm]{03/supervised/5_1} & \includegraphics[width=1.5cm, height=1.5cm]{03/supervised/6_1} & \includegraphics[width=1.5cm, height=1.5cm]{03/supervised/7_1} & & \includegraphics[width=1.5cm, height=1.5cm]{03/lci/1_0} & \includegraphics[width=1.5cm, height=1.5cm]{03/lci/2_0} & \includegraphics[width=1.5cm, height=1.5cm]{03/lci/3_0} & \includegraphics[width=1.5cm, height=1.5cm]{03/lci/4_0} & \includegraphics[width=1.5cm, height=1.5cm]{03/lci/5_0} & \includegraphics[width=1.5cm, height=1.5cm]{03/lci/6_1} & \includegraphics[width=1.5cm, height=1.5cm]{03/lci/7_0} & & \includegraphics[width=1.5cm, height=1.5cm]{03/ours/1_0} & \includegraphics[width=1.5cm, height=1.5cm]{03/ours/2_0} & \includegraphics[width=1.5cm, height=1.5cm]{03/ours/3_0} & \includegraphics[width=1.5cm, height=1.5cm]{03/ours/4_0} & \includegraphics[width=1.5cm, height=1.5cm]{03/ours/5_1} & \includegraphics[width=1.5cm, height=1.5cm]{03/ours/6_0} & \includegraphics[width=1.5cm, height=1.5cm]{03/ours/7_0}\\
% & & \cmark & \cmark & \cmark & \cmark & \cmark & \cmark & \cmark & & \cmark & \cmark & \cmark & \cmark & \cmark & \cmark & \cmark & & \cmark & \cmark & \cmark & \cmark & \cmark & \cmark & \cmark\\
\includegraphics[width=1.5cm, height=1.5cm]{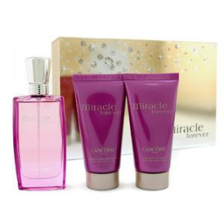} & & \includegraphics[width=1.5cm, height=1.5cm]{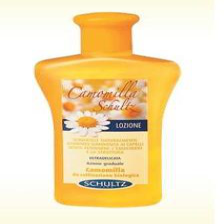} & \includegraphics[width=1.5cm, height=1.5cm]{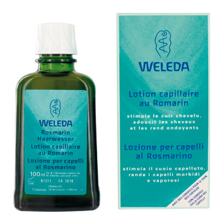} & \includegraphics[width=1.5cm, height=1.5cm]{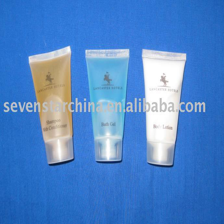} & \includegraphics[width=1.5cm, height=1.5cm]{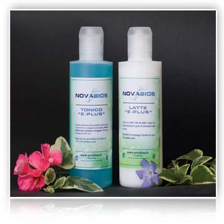} & \includegraphics[width=1.5cm, height=1.5cm]{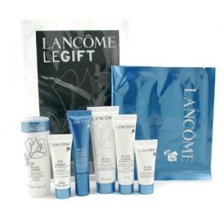} & \includegraphics[width=1.5cm, height=1.5cm]{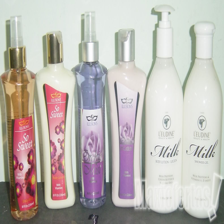} & \includegraphics[width=1.5cm, height=1.5cm]{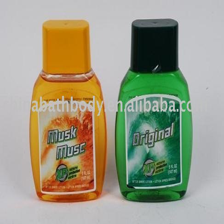} & & \includegraphics[width=1.5cm, height=1.5cm]{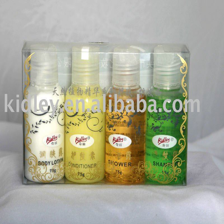} & \includegraphics[width=1.5cm, height=1.5cm]{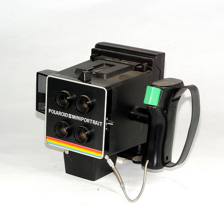} & \includegraphics[width=1.5cm, height=1.5cm]{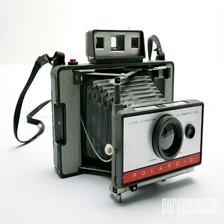} & \includegraphics[width=1.5cm, height=1.5cm]{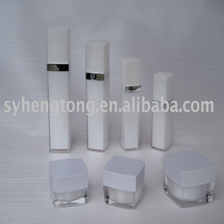} & \includegraphics[width=1.5cm, height=1.5cm]{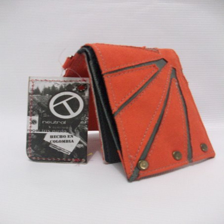} & \includegraphics[width=1.5cm, height=1.5cm]{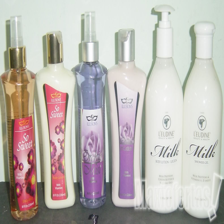} & \includegraphics[width=1.5cm, height=1.5cm]{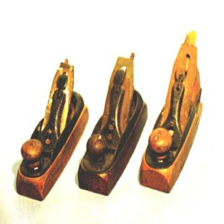} & & \includegraphics[width=1.5cm, height=1.5cm]{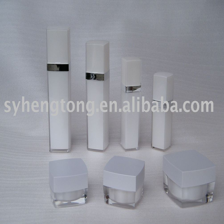} & \includegraphics[width=1.5cm, height=1.5cm]{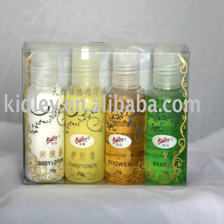} & \includegraphics[width=1.5cm, height=1.5cm]{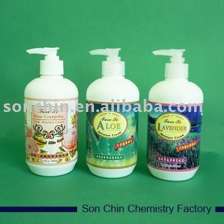} & \includegraphics[width=1.5cm, height=1.5cm]{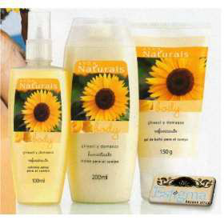} & \includegraphics[width=1.5cm, height=1.5cm]{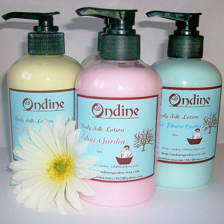} & \includegraphics[width=1.5cm, height=1.5cm]{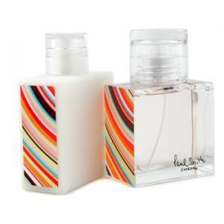} & \includegraphics[width=1.5cm, height=1.5cm]{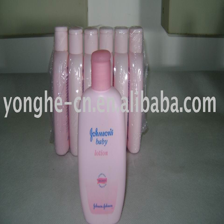}\\
& & \cmark & \cmark & \cmark & \cmark & \cmark & \cmark & \cmark & & \cmark & \xmark & \xmark & \cmark & \xmark & \cmark & \xmark & & \cmark & \cmark & \cmark & \cmark & \cmark & \cmark & \cmark\\
\includegraphics[width=1.5cm, height=1.5cm]{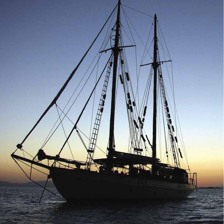} & & \includegraphics[width=1.5cm, height=1.5cm]{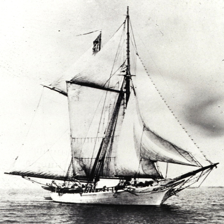} & \includegraphics[width=1.5cm, height=1.5cm]{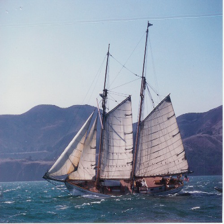} & \includegraphics[width=1.5cm, height=1.5cm]{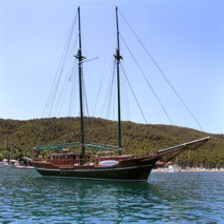} & \includegraphics[width=1.5cm, height=1.5cm]{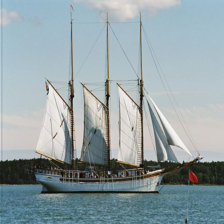} & \includegraphics[width=1.5cm, height=1.5cm]{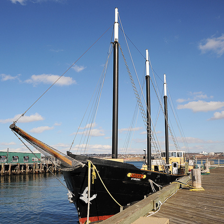} & \includegraphics[width=1.5cm, height=1.5cm]{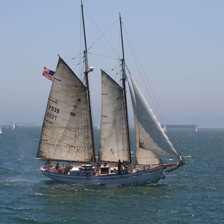} & \includegraphics[width=1.5cm, height=1.5cm]{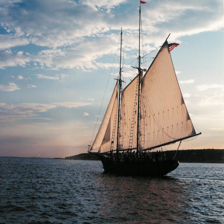} & & \includegraphics[width=1.5cm, height=1.5cm]{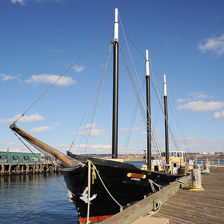} & \includegraphics[width=1.5cm, height=1.5cm]{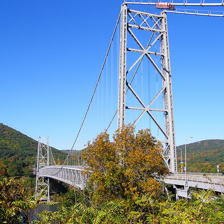} & \includegraphics[width=1.5cm, height=1.5cm]{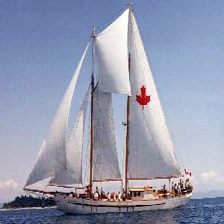} & \includegraphics[width=1.5cm, height=1.5cm]{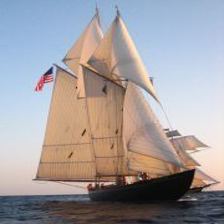} & \includegraphics[width=1.5cm, height=1.5cm]{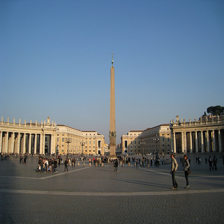} & \includegraphics[width=1.5cm, height=1.5cm]{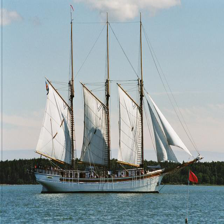} & \includegraphics[width=1.5cm, height=1.5cm]{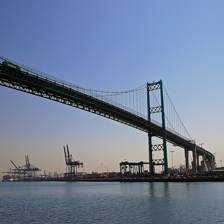} & & \includegraphics[width=1.5cm, height=1.5cm]{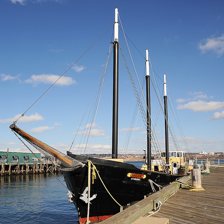} & \includegraphics[width=1.5cm, height=1.5cm]{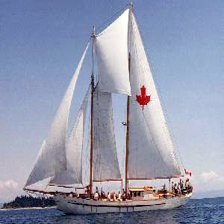} & \includegraphics[width=1.5cm, height=1.5cm]{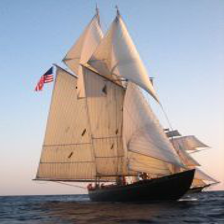} & \includegraphics[width=1.5cm, height=1.5cm]{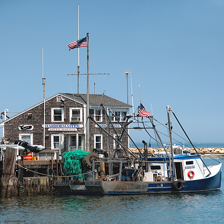} & \includegraphics[width=1.5cm, height=1.5cm]{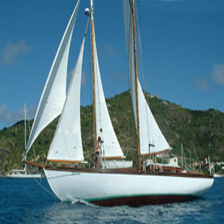} & \includegraphics[width=1.5cm, height=1.5cm]{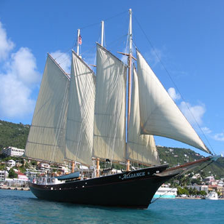} & \includegraphics[width=1.5cm, height=1.5cm]{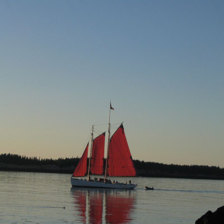}\\
& & \cmark & \cmark & \cmark & \cmark & \cmark & \cmark & \cmark & & \cmark & \xmark & \cmark & \cmark & \xmark & \cmark & \xmark & & \cmark & \cmark & \cmark & \cmark & \cmark & \cmark & \cmark\\
\includegraphics[width=1.5cm, height=1.5cm]{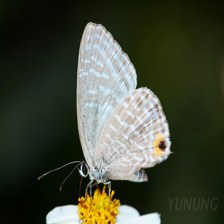} & & \includegraphics[width=1.5cm, height=1.5cm]{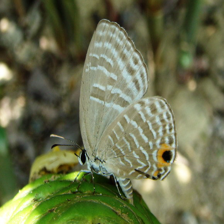} & \includegraphics[width=1.5cm, height=1.5cm]{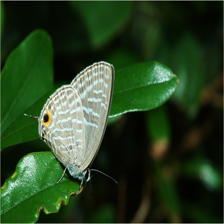} & \includegraphics[width=1.5cm, height=1.5cm]{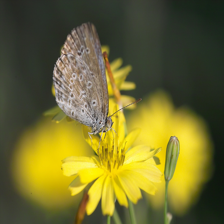} & \includegraphics[width=1.5cm, height=1.5cm]{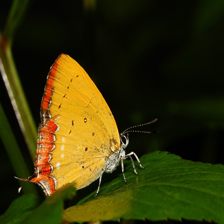} & \includegraphics[width=1.5cm, height=1.5cm]{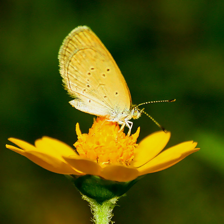} & \includegraphics[width=1.5cm, height=1.5cm]{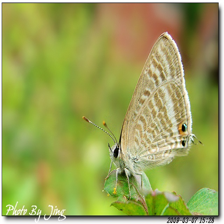} & \includegraphics[width=1.5cm, height=1.5cm]{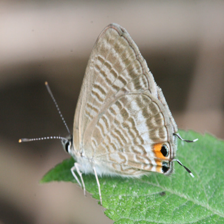} & & \includegraphics[width=1.5cm, height=1.5cm]{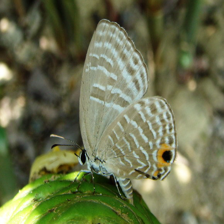} & \includegraphics[width=1.5cm, height=1.5cm]{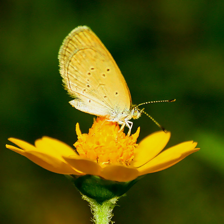} & \includegraphics[width=1.5cm, height=1.5cm]{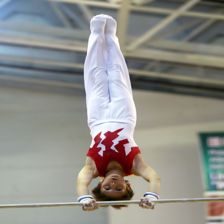} & \includegraphics[width=1.5cm, height=1.5cm]{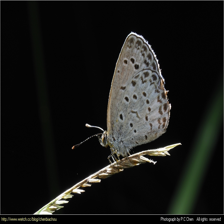} & \includegraphics[width=1.5cm, height=1.5cm]{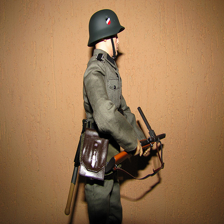} & \includegraphics[width=1.5cm, height=1.5cm]{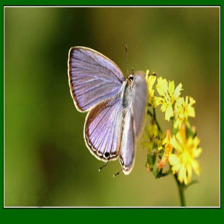} & \includegraphics[width=1.5cm, height=1.5cm]{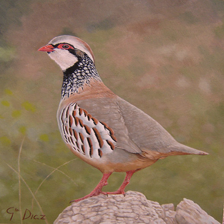} & & \includegraphics[width=1.5cm, height=1.5cm]{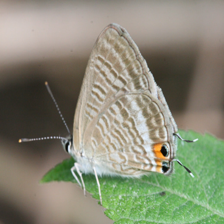} & \includegraphics[width=1.5cm, height=1.5cm]{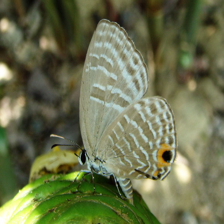} & \includegraphics[width=1.5cm, height=1.5cm]{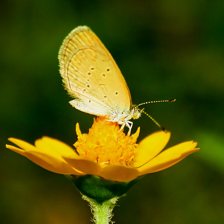} & \includegraphics[width=1.5cm, height=1.5cm]{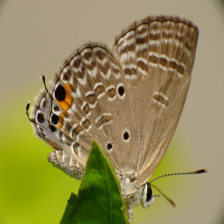} & \includegraphics[width=1.5cm, height=1.5cm]{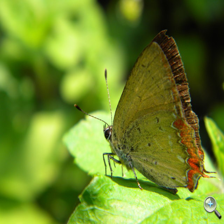} & \includegraphics[width=1.5cm, height=1.5cm]{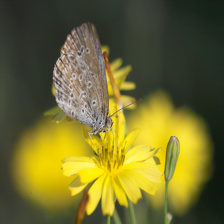} & \includegraphics[width=1.5cm, height=1.5cm]{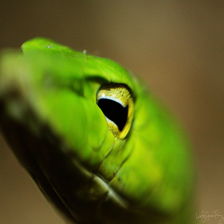}\\
& & \cmark & \cmark & \cmark & \cmark & \cmark & \cmark & \cmark & & \cmark & \cmark & \xmark & \cmark & \xmark & \cmark & \xmark & & \cmark & \cmark & \cmark & \cmark & \cmark & \cmark & \xmark
% \includegraphics[width=1.5cm, height=1.5cm]{07/query} & & \includegraphics[width=1.5cm, height=1.5cm]{07/supervised/1_1} & \includegraphics[width=1.5cm, height=1.5cm]{07/supervised/2_1} & \includegraphics[width=1.5cm, height=1.5cm]{07/supervised/3_1} & \includegraphics[width=1.5cm, height=1.5cm]{07/supervised/4_1} & \includegraphics[width=1.5cm, height=1.5cm]{07/supervised/5_1} & \includegraphics[width=1.5cm, height=1.5cm]{07/supervised/6_1} & \includegraphics[width=1.5cm, height=1.5cm]{07/supervised/7_1} & & \includegraphics[width=1.5cm, height=1.5cm]{07/lci/1_1} & \includegraphics[width=1.5cm, height=1.5cm]{07/lci/2_1} & \includegraphics[width=1.5cm, height=1.5cm]{07/lci/3_1} & \includegraphics[width=1.5cm, height=1.5cm]{07/lci/4_1} & \includegraphics[width=1.5cm, height=1.5cm]{07/lci/5_0} & \includegraphics[width=1.5cm, height=1.5cm]{07/lci/6_1} & \includegraphics[width=1.5cm, height=1.5cm]{07/lci/7_1} & & \includegraphics[width=1.5cm, height=1.5cm]{07/ours/1_1} & \includegraphics[width=1.5cm, height=1.5cm]{07/ours/2_1} & \includegraphics[width=1.5cm, height=1.5cm]{07/ours/3_1} & \includegraphics[width=1.5cm, height=1.5cm]{07/ours/4_1} & \includegraphics[width=1.5cm, height=1.5cm]{07/ours/5_1} & \includegraphics[width=1.5cm, height=1.5cm]{07/ours/6_0} & \includegraphics[width=1.5cm, height=1.5cm]{07/ours/7_1}\\
% & & \cmark & \cmark & \cmark & \cmark & \cmark & \cmark & \cmark & & \cmark & \cmark & \cmark & \cmark & \xmark & \cmark & \cmark & & \cmark & \cmark & \cmark & \cmark & \cmark & \cmark & \cmark\\
\end{tabular}}
\end{center}
\caption{Comparison of nearest neighbor retrieval results. The left most column shows the query images. In each of the three consecutive columns, we show the seven nearest neighbors of the query image with features respectively learned using ImageNet labels, GlobStat \cite{Jenni2020GlobStat} and our model CODIAL. Queries contain randomly selected images possessing variant characteristics. Semantically related and unrelated retrievals are respectively marked with green tick and red cross. (best viewed in color)}
\vspace{-10pt}
\label{fig:qual_results_nnr}
\end{figure*}

\myparagraph{Transfer Learning on PASCAL VOC:}
We test the transferability of the learned feature on PASCAL VOC dataset \cite{Everingham2014PASCAL}. We use our unsupervised trained network $F_\theta$ as the initialization model for variant tasks, such as multi-label image classification, object detection and semantic segmentation. Performance is measured by mean average precision (mAP) for classification and detection, and by mean intersection over union (mIoU) for segmentation task. We follow the established setup of \cite{Krahenbuhl2016DDI} for multi-label classification, the Fast-RCNN framework \cite{Girshick2015FRCNN} for detection and the FCN framework \cite{Shelhamer2016FCN} for semantic segmentation. For classification, we extract the \texttt{conv5} features on the PASCAL VOC 2007 images and train a regularized multinomial logistic regression classifier on top of those frozen representation by minimizing the cross-entropy objective using LBFGS with $\ell_2$-regularization with standard parameters. In the case of object detection, the self-supervised weights learned by our model are used to initialize the Fast-RCNN model \cite{Girshick2015FRCNN} and use a multi-scale training and single-scale testing on the PASCAL VOC 2007 images. For segmentation, we fine-tune the self-supervised features learned by our model using FCN \cite{Shelhamer2016FCN} on PASCAL VOC 2012.

\tab{tab:imagenet_results} (right) summarizes the comparison of our approach with other methods. We consistently outperform previous methods on all these three tasks, particularly on classification, detection and segmentation tasks, we respectively achieve 0.7\%, 0.8\% and 0.2\% margin gain compared to the challenging prior arts, which is quite remarkable and can be seen as the advantage of our hybrid method that jointly solve the discriminative and aligning tasks.

% \massi{(M: One thing of this part is the fact that implementation details surpass the comment on the results. While it is true that the comment cannot differ much from the previous, is there something we can add? For the implementation details, maybe we could cite some works (to avoid reporting details) and/or move something to the supplementary.)}

\subsection{Qualitative Results}

Self-supervised training associates similar features to semantically similar images. In this section, we visually investigate whether our features fulfill this property. Additionally, we qualitatively compare our results with the features learned by supervised `ImageNet Labels' model and the recent self-supervised model GlobStat \cite{Jenni2020GlobStat}. For doing so, we compute the nearest neighbors of the SSL (for model proposed by GlobStat \cite{Jenni2020GlobStat} and us) and SL (model trained with `ImageNet Labels') features of \texttt{conv5} layers of AlexNet on the validation set of ImageNet-1K. For our model, we obtain features from the 4,096 dimensional vector outputted by the feature extractor network $F_\theta$. We use cosine similarity to calculate the distance between features.

The images are arranged from left to right in order of increasing distance in \fig{fig:qual_results_nnr}. We observe that all the models are able to capture semantic information in images for some queries. In some cases, our retrievals are quite similar to the ones returned by GlobStat \cite{Jenni2020GlobStat}, however, many images retrieved by \cite{Jenni2020GlobStat} are unrelated to the query. This is likely because GlobStat \cite{Jenni2020GlobStat} focuses more on the texture and shape of object and is less discriminative towards different instances. As a result, it retrieves many instances based on the background information, such as in rows 1 and 4, some wrongly retrieved images are found based on the texture of sky. In row 2, background and shape of the query animal influence the wrong retrieval. Also in row 3, some wrong examples are retrieved based on the white background, we speculate that happens because of its ability to focus on texture information. On the contrary, our model can return more semantically similar images for these queries, which confirms our model’s discriminative ability on instance level as a benefit to the maximization of mutual information.
% \massi{(M: I found this sentence a bit unclear. In particular, if LCI focuses on the shape, should not be able to retrieve instances having similar shape and likely of the same semantic concepts (or, at least, related ones?). Looking at the mid row, it seems it retrieved some instances based on the white background also.)}

% compare the learned features of our model and the ones. We first perform nearest-neighbor retrieval on ImageNet-1K validation set to test the ability of learned features in capturing semantic meanings. We compare our self-supervised features with the supervised features learned with labeled supervision. For our model, we obtain features from the 4,096 dimensional vector outputted by the feature extractor network $F_\theta(\cdot)$.

% Conclusions
\section{Conclusion}
\label{sec:concl}
We presented CODIAL, a self-supervised visual representation learning method by jointly solving discriminating and aligning tasks. The discriminative task involves recognizing image transformations, such as rotation angle, warping, that needs learning global statistics of the image, whereas the alignment is done by maximizing the mutual information between the transformed version of the same image. This principle explicitly describes which features to be closed together or separated. We present the efficacy of our proposal through substantial experiments on nine self-supervised and transfer learning benchmark datasets where our model consistently outperforms the existing methods in various settings and tasks.

% Acknowledgment
\section*{Acknowledgments}
This work has been partially funded by the ERC starting grant DEXIM (grant agreement no. 853489) and DFG-EXC-Nummer 2064/1-Projektnummer 390727645.

{\small
\bibliographystyle{ieee_fullname}
\bibliography{bibliography}
}

% Appendix
\section*{Appendix}
\myparagraph{Experiments with $\beta$ on STL-10:} Since our mutual information estimator is trained together with other parametric models, the hyperparameter $\beta$ is slowly increased during training, starting from a very small value $10^{-6}$ to its final value with an exponential scheduling. We set the number of epochs by which $\beta$ reaches to $\beta_\text{end\_value}$ from $\beta_\text{start\_value}$ equal to 100. We experiment with two parameters $\beta_\text{start\_epoch}$ and $\beta_\text{end\_value}$ which are related to the $\beta$ hyperparameter. First, we experiment with the start epoch ($\beta_\text{start\_epoch}$) which indicates the epoch at which $\beta$ starts increasing to the final value $\beta_\text{end\_value}=1.0$ during 100 epochs. We consider $\beta_\text{start\_epoch}=10,30,50,70,90$ and the obtained results are shown in \tab{tab:stl10_beta} (top), where we observe that for $\beta_\text{start\_epoch}=10$, the obtained results across all the convolutional layers of AlexNet are consistently better that the other $\beta_\text{start\_epoch}$s. Next, we experiment with $\beta_\text{end\_value}=0.00001,0.0001,0.01,1.0$ and set $\beta_\text{start\_epoch}=10$ (as revealed the best), and the obtained results are presented in \tab{tab:stl10_beta} (bottom), where we can see that with $\beta_\text{end\_value}=1.0$, we steadily achieve superior results across different convolutional layers.

{
\setlength{\tabcolsep}{2.5pt}
\renewcommand{\arraystretch}{1.2} 
\begin{table}[!ht]
\centering
\begin{tabular}{l|ccccc}
 & \multicolumn{5}{c}{ \textbf{STL-10}} \\
 \hline
$\beta_\text{start\_epoch}$ & \texttt{c1} & \texttt{c2} & \texttt{c3} & \texttt{c4} & \texttt{c5}\\
\hline
10 & \textbf{60.5} & 71.5 & \textbf{74.3} & \textbf{75.3} & 75.4\\
30 & 59.7 &	70.7 & 73.6 & 74.7 & 75.2\\
50 & 59.6 & \textbf{71.9} & 74.0 & 74.5 & 74.4\\
70 & 59.8 & 70.4 & 74.0 & 74.6 & 75.3\\
90 & 60.3 & 71.3 & 73.4 & 74.6 & \textbf{75.5}\\
\hline
$\beta_\text{end\_value}$ & \texttt{c1} & \texttt{c2} & \texttt{c3} & \texttt{c4} & \texttt{c5}\\
\hline
0.00001 & 60.1 & 71.3 & 73.2 & 74.4 & 74.7\\
0.0001 & 60.3 & 71.3 & 73.6 & 74.6 & 74.9\\
0.01 & \textbf{60.5} & \textbf{71.6} & 73.7 & 75.1 & 75.1\\
1.0 & \textbf{60.5} & 71.5 & \textbf{74.3} & \textbf{75.3} & \textbf{75.4}
\end{tabular}
\caption{Ablating different $\beta_\text{start\_epoch}$s (top): We report test set performance of our pre-text model trained with different $\beta_\text{start\_epoch}$. Ablating different $\beta_\text{end\_value}$s (bottom): We report the test set performance of our final model by varying the $\beta_\text{end\_value}$s. (STL-10 with CNN backbone AlexNet with conv layers (\texttt{c1-5}).}
\label{tab:stl10_beta}
\end{table}
}

\myparagraph{ResNet Experiments on STL-10:} We also perform additional experiments with a more modern network architecture on STL-10. For doing so, we follow \cite{Ji2019IIC,Jenni2020GlobStat} and consider the ResNet-34 \cite{He2016ResNet} framework instead of the AlexNet \cite{Krizhevsky2012AlexNet}. We train our model to solve our hybrid discriminating and aligning pretext task for 200 epochs on 100K unlabeled training samples of STL-10. Once pretrained, we use those weights to initialize the network for downstream classification task on STL-10, and fine tune the model for 300 epochs on the 5K labeled training images and evaluate on the 8K test images.

\begin{table}[!ht]
\centering
\begin{tabular}{lc}
\toprule
\textbf{Method} & \textbf{Accuracy} \\
\midrule
MultTaskBayes \cite{Swersky2013MTBayesian} & 70.1\%\\
DiscUFL \cite{Dosovitskiy2015CNN} & 74.2\%\\
StackedAE \cite{Zhao2016StackedAE} & 74.3\%\\
DiscAttr \cite{Huang2016DiscAttr} & 76.8\%\\
ScaleScatter \cite{Oyallon2017ScaleScatter} & 87.6\%\\
SpotArtifacts \cite{Jenni2018SSLArtifacts} & 80.1\%\\
InfoMax \cite{Hjelm2019InfoMax} & 77.0\%\\
IIC \cite{Ji2019IIC} & 88.8\%\\
GlobStat \cite{Jenni2020GlobStat} & 91.8\%\\
\midrule
Ours & \textbf{92.1\%}\\
\bottomrule
\end{tabular}
\caption{Comparison of test set accuracy on STL-10 with other published results.}
\label{tab:results_resnet34}
\end{table}

We compare our CODIAL model with nine existing works. Among them, MultTaskBayes \cite{Swersky2013MTBayesian} proposes multi-task Gaussian processes to the Bayesian optimization framework; DiscUFL \cite{Dosovitskiy2015CNN} trains the network to discriminate between a set of surrogate classes; StackedAE \cite{Zhao2016StackedAE} presents an architecture, called \emph{stacked what-where auto-encoders}, which integrates discriminative and generative pathways and uses a convolutional net to encode the input, and employs a deconvolutional net to produce the reconstruction; DiscAttr \cite{Huang2016DiscAttr} trains a CNN coupled with unsupervised discriminative clustering, which uses the cluster membership as a soft supervision to discover shared attributes from the clusters while maximizing their separability; ScaleScatter \cite{Oyallon2017ScaleScatter} uses the scattering transform in combination with convolutional architectures; SpotArtifacts \cite{Jenni2018SSLArtifacts} learns self-supervised knowledge by spotting synthetic artifacts in images; InfoMax \cite{Hjelm2019InfoMax} learns representations by maximizing mutual information between two transformed views of the same image which are formed by applying a variety of transformations to a randomly sampled `seed’ image patch; and GlobStat \cite{Jenni2020GlobStat} distinguishes diverse image transformations, such as rotation angles, warping and limited context inpainting. \tab{tab:results_resnet34} presents the results obtained by our CODIAL model and compares it with other methods mentioned above, which shows that our CODIAL achieves highest results with ResNet-34 backbone as well and surpasses the closest model GlobStat by a margin of 0.3\%. This proves that our model is effective with other backbone network as well and can be benefited from our joint discriminating and aligning pretext task.

\end{document}